\documentclass[11pt]{article}

\usepackage[]{acl}
\usepackage{multirow}
\usepackage{times}
\usepackage{booktabs}

\usepackage{latexsym}
\usepackage{amsmath}
\usepackage{amssymb}
\usepackage{tikz}
\usetikzlibrary{arrows.meta, positioning}
\usepackage{comment}
\usepackage[T1]{fontenc}

\usepackage[utf8]{inputenc}

\usepackage{microtype}

\usepackage{inconsolata}

\usepackage{graphicx}
\graphicspath{{figs/}{}}
\usepackage{listings}
\usepackage{xcolor}

\definecolor{codegray}{rgb}{0.95,0.95,0.95}
\definecolor{codeblue}{rgb}{0.2,0.2,0.7}
\definecolor{codegreen}{rgb}{0,0.5,0}
\definecolor{codered}{rgb}{0.7,0.1,0.1}

\lstdefinestyle{mystyle}{
    backgroundcolor=\color{codegray},
    commentstyle=\color{codegreen},
    keywordstyle=\color{codeblue},
    stringstyle=\color{codered},
    basicstyle=\ttfamily\small,
    breakatwhitespace=false,
    breaklines=true,
    captionpos=b,
    keepspaces=true,
    numbers=none,
    showspaces=false,
    showstringspaces=false,
    showtabs=false,
    tabsize=2,
    frame=single,
    frameround=tttt
}

\title{QuadAI at SemEval-2026 Task 3: \\Ensemble Learning of Hybrid RoBERTa and LLMs for Dimensional Aspect-Based Sentiment Analysis}

\author{
           A.J.W. de Vink$^{1}$, Filippos Karolos Ventirozos$^{2}$, Natalia Amat-Lefort$^{1}$, Lifeng Han$^{*, 1,3}$
              \vspace*{0.075cm}
\\            $^1$ LIACS, Leiden University, NL
\\            $^2$ Manchester Metropolitan University, UK
  \\          $^3$  Leiden University Medical Center, NL  
      \\   $^*$\textit{Corresponding Author: l.han@liacs.leidenuniv.nl} 
}

\begin{document}
\maketitle
\begin{abstract}
We present our system for SemEval-2026 Task 3 on dimensional aspect-based sentiment regression. Our approach combines a hybrid RoBERTa encoder, which jointly predicts sentiment using regression and discretized classification heads, with large language models (LLMs) via prediction-level ensemble learning. The hybrid encoder improves prediction stability by combining continuous and discretized sentiment representations. We further explore in-context learning with LLMs and ridge-regression stacking to combine encoder and LLM predictions. Experimental results on the development set show that ensemble learning significantly improves performance over individual models, achieving substantial reductions in RMSE and improvements in correlation scores. Our findings demonstrate the complementary strengths of encoder-based and LLM-based approaches for dimensional sentiment analysis.
Our development code and resources will be shared at \url{https://github.com/aaronlifenghan/ABSentiment}.

\end{abstract}

\section{Introduction}
\label{sec:intro}

Aspect-based sentiment analysis (ABSA) is a natural language processing (NLP) task that involves a few sub-tasks that include aspect term extraction, aspect category detection, opinion extraction, and aspect sentiment classification \cite{zhang2022survey}. 
It has witnessed the traditional ML, deep learning, and pretrained language model (PLM) approaches. 
The Transformer-based models (both encoders \cite{liao2021improved,chauhan2025enhanced} and decoders \cite{mughal2024comparative,ventirozos-etal-2025-aspect,ventirozos-etal-2025-sure}) have been the dominant methods nowadays for such tasks, while challenges remain, such as data scarcity, domain application, and modeling complex aspect-opinion relationships \cite{nazir2020issues,zhang2022survey}.


The shared task we attended this year has two tracks: Track-A ``Dimensional Aspect-Based Sentiment Analysis (DimABSA)'' and Track-B ``Dimensional Stance Analysis (DimStance)''.
We attended Trask-A.1 ``DimASR - Valence-Arousal (VA) Prediction'' and Track-B, leaving ``A.2: DimASTE - Triplet Extraction'' and ``A.3: DimASQP - Quadruplet Extraction'' into our future work.
The shared task data is described at \cite{lee2026dimabsabuildingmultilingualmultidomain,becker2026dimstancemultilingualdatasetsdimensional}
As examples of Task-A1 and Task-B, we list the text format of input and output in Figure \ref{fig:example_taskA1} and \ref{fig:example_taskB1}.
Valence indicates positivity and negativity with more dimensions.
Arousal indicates emotional intensity from high to low. 
For instance, (Happy, Delight, Excited) can be located in the corner of ``positive and high'' emotion, while (depressed, bored, tired) can be in the ``negative and low'' emotion\footnote{ \url{https://github.com/DimABSA/DimABSA2026}}.

In this work, we introduce related work to our proposed method, the methodology design and development (Hybrid RoBERTa, LLMs, ensemble), the evaluation results from development sets, and our submissions to the shared task. Due to unforeseen circumstances, we did not manage to submit all the methods we developed; however, we will continue our testing for offline development, and our codes and resources will be shared publicly for open science.

\section{Related Work}
\label{sec:related}



\subsection{Beyond VA Scores}
In addition to the Valance and Arousal (VA) score, 
\newcite{shi2025triple} tried to use another psychological emotion dimension ``Dominance'', i.e., the degree of control or influence.
To address the issues of standard ABSA models that rely on word embeddings and attention but do not use structured emotion knowledge, they proposed the Graph Attention Network (GAT) method to model relationships between words and capture the syntactic dependencies. Similarly, there are other recent works using graph knowledge to address sentiment analysis, such as ReviewGraph \cite{de2025reviewgraph}.


\subsection{Language/Domain Specific ABSA}
For language-specific ABSA, \newcite{lee-etal-2024-overview-sighan} introduced a shared task for the Chinese language, which attracted 11 teams, focusing on intensity prediction, triplet extraction and quadruple extraction.

For domain specific work on ABSA, \newcite{chakraborty2020aspect} used active learning on scientific reviews, using ~8,000 peer reviews from ICLR conference, including (review text, score, paper decision). The work focused on the relationship between aspect sentiment (positive/negative) and paper outcome (accept/reject).

\subsection{Hybrid Models for ABSA}
\newcite{zhang2024hybrid} used hybrid setting of BERT based encoder models and LLMs.
They first use BERT pipeline to extract aspects, categories, and opinions, then use LLM with QLoRA fine-tuning to predict sentiment intensity based on the BERT outputs.


Another hybrid model conducted by \newcite{liang2022aspect} combines neural networks and contextual feature representations. Their model integrates word embeddings with attention mechanisms to capture aspect-specific contextual information.
Experimental results demonstrated that the hybrid approach outperforms traditional neural models on benchmark datasets. This work highlights the effectiveness of hybrid architectures for fine-grained sentiment prediction.

More challenges, tasks, and methodologies on ABSA can be found in earlier surveys \cite{nazir2020issues,zhang2022survey}.




\section{Methodology}
\label{sec:method}

\begin{figure}[t]
\centering 
\includegraphics[width=0.4\textwidth]{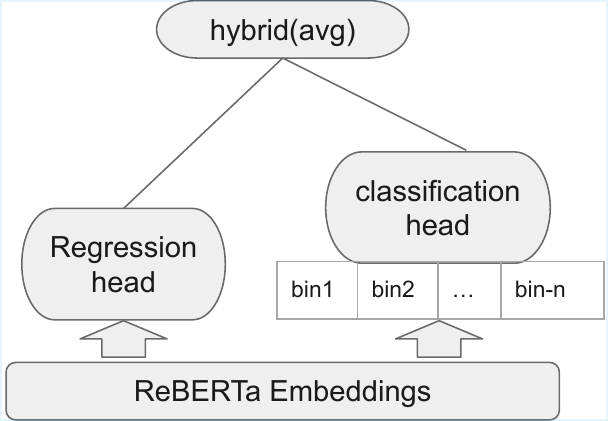}
  \caption{Hybrid RoBERTa
  } 
  \label{fig:hybrid_RoBERTa}
\end{figure}
\subsection{Hybrid RoBERTa}
We designed a hybrid encoder-based model using averaged scores from regression and discretized bin integrated classification, as showin in Figure \ref{fig:hybrid_RoBERTa}. 
We firstly used RoBERTa embedding as the encoder.
Then, in parallel, we trained a regression head and a discretized classification head. 
The core idea of discretizing the target space is that we take the continuous embedding variable and split it into n bins. The final layer outputs an n-dimensional logit vector then applied with softmax and trained with cross-entropy loss.
The advantage of discretized classification is that it is expected to be a more stable training than regression and expresses confidence over bins.

Lastly, the final prediction of Hybrid RoBERTa is obtained by averaging both outputs ($w=0.5$). 
We list the equations below.
Regression output:

\begin{equation}
\hat{y}_{reg}
\end{equation}

Classification expected value:

\begin{equation}
\hat{y}_{cls} =
\sum_{i=1}^{B}
p_i c_i
\end{equation}

Final prediction:

\begin{equation}
\hat{y}
=
w \hat{y}_{reg}
+
(1-w) \hat{y}_{cls}
\end{equation}

Training objective:

\begin{equation}
L =
L_{reg}
+
\alpha L_{cls}
\end{equation}

\begin{figure}[t]
\centering   \includegraphics[width=0.35\textwidth]{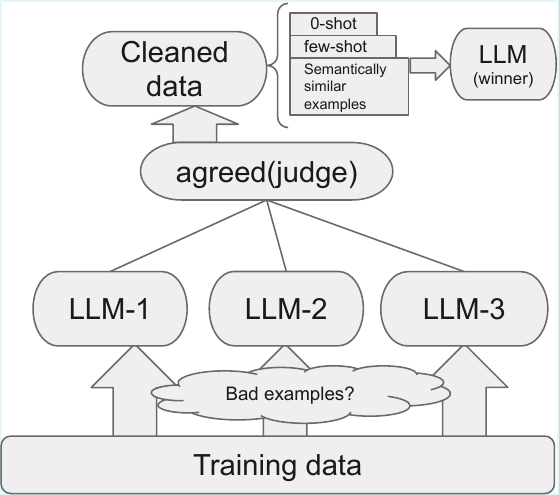}
  \caption{Triple-LLMs workflow 
  } 
  \label{fig:3LLMs}
\end{figure}

\subsection{LLMs}

As a starting point, we explore the difference between:
\begin{itemize}
    \item zero-shot prompting (no examples)
    \item random examples (40, 60)
    \item semantically picking similar examples (40, 60, 100, 200, 600)
\end{itemize}
The semantical similarity is according to the embedding similarity scores at sentence level. We utilized the OpenAI's model for sentence embeddings\footnote{text-embedding-3-large}.

Looking into some examples, we also decided to filter out low quality ones, such as cases where the labels do not look right. We call this step as Data Cleaning.
The step involved firstly using HDBScan\footnote{\url{https://github.com/scikit-learn-contrib/hdbscan}} to cluster all the ones used for in-context learning (i.e. training-set) instances according to the two dimensions of valence and arousal. HDBScan uses the DBScan approach but converts it into hierarchical clustering, for which we used the given, default, hyper-parameters for the clustering. 

Following for each cluster we had separate three LLMs be presented each cluster, similar VA scores, and ask in the prompt to pinpoint which one of these is an outlier or not. The prompt for that can be found under Appendix~\ref{sec:appendix:prompt}. Finally, if all three LLMs agree that a specific instance(s) in a cluster is at outlier, we would remove them from the available pool of in-context learning candidates.



This is shown in Figure \ref{fig:3LLMs}.
Exact LLMs we used for this task are ``gemini'', ``claude'', and ``gpt5.2'' for data cleaning with cross validation, and ``gemini'' for last stage prompting as the best performing LLM.





To reduce variance and improve model robustness, we explore the ensemble learning strategy, which will be described below. 
\subsection{Ensemble Learning}
\begin{figure}[t]
\centering
\resizebox{\columnwidth}{!}{
\begin{tikzpicture}[
    node distance=1.2cm and 1.2cm,
    box/.style={draw, rounded corners, minimum width=2.0cm, minimum height=0.7cm, align=center, font=\small},
    smallbox/.style={draw, rounded corners, minimum width=1.8cm, minimum height=0.6cm, align=center, font=\small},
    arrow/.style={-Latex, thick}
]

\node[box] (text) {Input Text};

\node[smallbox, below left=of text] (m1) {Model$_1$\\$\hat{y}_1$};
\node[smallbox, below right=of text] (m2) {Model$_2$\\$\hat{y}_2$};

\node[smallbox, below=of text] (vader) {VADER\\features};

\node[box, below=2.2cm of text] (comb) {Combiner\\(avg / weighted / ridge)};

\node[box, below=of comb] (out) {Final prediction\\$\hat{y}$};

\draw[arrow] (text) -- (m1);
\draw[arrow] (text) -- (m2);
\draw[arrow] (text) -- (vader);

\draw[arrow] (m1) -- (comb);
\draw[arrow] (m2) -- (comb);
\draw[arrow] (vader) -- (comb);

\draw[arrow] (comb) -- (out);

\end{tikzpicture}
}
\caption{Prediction-level ensemble architecture combining base models and optional VADER features.}
\label{fig:ensemble-diag}
\end{figure}
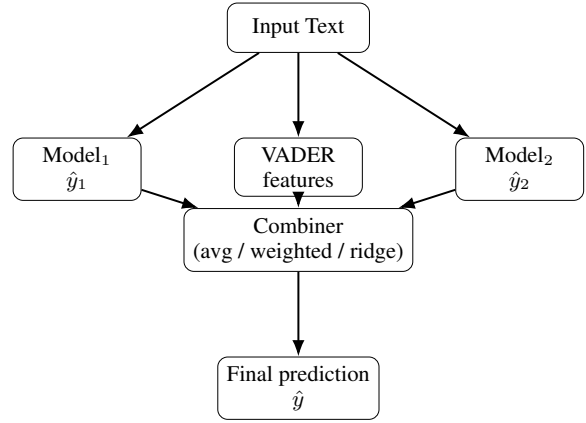

For Ensemble Learning, we design the prediction-level fusion (\textit{aka} late fusion or model stacking) of Hybrid RoBERTa and LLMs with optional other features, as in Figure \ref{fig:ensemble-diag}.
For this work, we incorporate lexical sentiment features derived from VADER \cite{hutto2014vader}, including compound, positive, negative, and neutral polarity scores, as auxiliary inputs to the ensemble combiner.











We detail the ensemble prediction into mathematical formulas below:
Given $K$ base models, each producing a prediction $\hat{y}_k$ for an input $x$,
the final prediction is obtained via a combiner $g(\cdot)$:

\begin{equation}
\hat{y} = g(\hat{y}_1, \hat{y}_2, \dots, \hat{y}_K, \mathbf{f}),
\end{equation}

where $\mathbf{f}$ denotes optional external features (e.g., VADER scores).

\textbf{Simple averaging:}
\begin{equation}
\hat{y} = \frac{1}{K} \sum_{k=1}^{K} \hat{y}_k
\end{equation}

\textbf{Weighted averaging:}
\begin{equation}
\hat{y} = \sum_{k=1}^{K} w_k \hat{y}_k,
\quad \sum_{k=1}^{K} w_k = 1
\end{equation}

\textbf{Ridge stacking:}
\begin{equation}
\hat{y} = \mathbf{w}^\top
[\hat{y}_1, \hat{y}_2, \dots, \hat{y}_K, \mathbf{f}] + b
\end{equation}


The VADER features include: compound  (continuous [-1,1]), pos, neu, neg. So the stacking vector becomes: 
\[
\mathbf{x} =
[\hat{y}_1,\ \hat{y}_2,\ \text{compound},\ \text{pos},\ \text{neu},\ \text{neg}]
\]
The rationale for including VADER as a feature is that the ensemble combines a lexicon and rule-based component, which is domain-robust and fast, with neural-based encoder models and LLMs.

\textbf{Training objective (ridge stacking).}
Given training targets $\mathbf{y}\in\mathbb{R}^n$ and the design matrix
$\mathbf{X}\in\mathbb{R}^{n\times (K+F)}$ whose $i$-th row is
$\mathbf{x}^{(i)}=[\hat{y}_1^{(i)},\ldots,\hat{y}_K^{(i)},\mathbf{f}^{(i)}]$,
we learn $\mathbf{w}$ and $b$ by:

\begin{equation}
\min_{\mathbf{w},\,b}\;
\frac{1}{n}\left\lVert \mathbf{y}-\left(\mathbf{X}\mathbf{w}+b\mathbf{1}\right)\right\rVert_2^2
\;+\;\lambda \lVert \mathbf{w} \rVert_2^2,
\end{equation}
% ...

We construct $\mathbf{X}$ using out-of-fold predictions of the base models to avoid label leakage.

\begin{table*}[t!]
\centering
\small
\caption{Performance of the Hybrid RoBERTa model on SemEval Task 1 (\textbf{Laptop}, Dev set). 
The model combines a regression head and a discretized classification head (31 bins), 
with the final prediction obtained by averaging both outputs ($w=0.5$). Error catergories: lower is better; in 3 catergories hyrid is much better, except for RMSE that is comparable to Bin. Pearson correation $\rho$ higher is better rho( Regress>Hybrid>Bin)
}
\label{tab:hybrid_main}
\begin{tabular}{lccccccc}
\hline
Model Variant & MSE $\downarrow$ & RMSE $\downarrow$ & RMSE$_v$ $\downarrow$ & RMSE$_a$ $\downarrow$ & $\rho_v$ $\uparrow$ & $\rho_a$ $\uparrow$ & $\rho_{mean}$ $\uparrow$ \\
\hline
Regression only & 0.6140 & 0.7836 & 0.7201 & 0.8423 & 0.9102 & 0.5505 & 0.7304 \\
Bin (expected value) & 0.6238 & 0.7898 & 0.8340 & \textbf{0.7430} & 0.8974 & 0.5126 & 0.7050 \\
Hybrid (average) & \textbf{0.5419} & \textbf{0.7361} & \textbf{0.7214} & {0.7506} & 0.9074 & {0.5388} & {0.7231} \\
\hline
\end{tabular}
\end{table*}

\begin{table}[t!]
\centering
\tiny 
\caption{Top-10 Hybrid configurations ranked by average RMSE on the \textbf{Laptop} Dev set. 
$num\_bins$ denotes the number of discretization bins, $\alpha$ the classification loss weight, 
and $w$ the regression--classification averaging weight. 
}
\label{tab:hybrid_sweep}
\begin{tabular}{cccccc}
\hline
$num\_bins$ & $\alpha$ & $w$ & RMSE$_{avg}$ $\downarrow$ & $\rho_{mean}$ $\uparrow$ & $\rho_a$ $\uparrow$ \\
\hline
31 & 0.5 & 0.5 & \textbf{0.7361} & 0.7231 & 0.5388 \\
11 & 1.0 & 0.5 & 0.7368 & 0.7182 & 0.5277 \\
31 & 0.5 & 0.4 & 0.7376 & 0.7252 & 0.5421 \\
7  & 0.2 & 0.5 & 0.7405 & 0.7287 & 0.5473 \\
31 & 0.5 & 0.3 & 0.7431 & 0.7270 & 0.5448 \\
11 & 1.0 & 0.4 & 0.7448 & 0.7185 & 0.5278 \\
7  & 0.2 & 0.4 & 0.7473 & 0.7293 & 0.5485 \\
11 & 0.5 & 0.5 & 0.7481 & 0.7251 & 0.5523 \\
31 & 1.0 & 0.5 & 0.7488 & 0.7138 & 0.5278 \\
31 & 0.2 & 0.4 & 0.7516 & 0.7151 & 0.5234 \\
\hline
\end{tabular}
\end{table}

\section{Model Training and Development}


\subsection{Hybrid ReBERTa for Track-A1}
The system performance on the development set of \textit{laptop} category from encoder-based models is shown in Table \ref{tab:hybrid_main} and \ref{tab:hybrid_sweep}. 
Table \ref{tab:hybrid_main} presents the overall comparisons among regression-only, discretization-bin, and hybrid (averaging two). From the Error score metrics, we can see that the Hybrid model produced much better output on two metrics, MSE and RMSE, which have a bigger margin decrease in the error scores. However, it produced similar (or comparable) scores on RMSE(v) and RMSE(a), to regression-only and Bin models, respectively.
For Pearson correlation scores, the Regression-only model produced the highest scores, though not much difference from the Hybrid model.
Table \ref{tab:hybrid_sweep} displays the Top-10 hybrid configurations ranked by averaging RMSE on the Laptop Dev set. We tried different sets of triple values for the parameters exhaustively; however, for future development, it would be more suitable to carry out automated hyperparameter tuning, e.g., OPTUNA \cite{akiba2019optunanextgenerationhyperparameteroptimization} to explore.

In addition, Table \ref{tab:restaurant_hybrid_results} shows the Hybrid RoBERTa performance on Task-A1 \textit{restaurant} dev data. We can see from the scores that the hybrid model achieved the best performance on this data across all tested metrics, including both error scores and correlations. Importantly, the MSE score of the hybrid model is almost down to half that of the regression model (0.4919 vs 0.8176). Large improvement margins can also be observed from RMSE scores.

\subsection{LLMs on Track-A Laptop Dev}
The LLM output evaluation on the \textit{laptop} Dev set is shown in Table \ref{tab:llm-dev}, where we can see that, in comparison to Hybrid RoBERT in Table \ref{tab:hybrid_sweep}, the LLMs produced an even lower RMSE score of 0.695, vs 0.7361.
In addition, it increased the correlation mean score from 0.7231 to 0.757. 









\begin{table*}[t!]
\centering
\small
\caption{\textbf{Hybrid} RoBERTa results on the \textbf{Restaurant} development set for SemEval Task~1 (Valence/Arousal). The model combines a regression head and a hard-bin classification head; the \emph{average} prediction is a weighted combination with $\texttt{pred\_weight}=0.5$. Best configuration and detailed dev metrics are shown.}
\label{tab:restaurant_hybrid_results}
\begin{tabular}{llccccccc}
\hline
Domain & Variant & MSE$\downarrow$ & RMSE$\downarrow$ & RMSE$_v$$\downarrow$ & RMSE$_a$$\downarrow$ & $\rho_v$$\uparrow$ & $\rho_a$$\uparrow$ & $\rho_{mean}$$\uparrow$ \\
\hline
\multirow{3}{*}{Restaurant}
& Regression & 0.8176 & 0.9042 & 0.8212 & 0.9802 & 0.9205 & 0.6458 & 0.7832 \\
& Bin-expected & 0.5369 & 0.7327 & 0.8154 & 0.6395 & 0.9130 & 0.6408 & 0.7769 \\
& Average ($w=0.5$) & \textbf{0.4919} & \textbf{0.7013} & \textbf{0.6692} & \textbf{0.7320} & \textbf{0.9217} & \textbf{0.6679} & \textbf{0.7948} \\
\hline
\end{tabular}

\vspace{2mm}
\footnotesize
Best config (Restaurant): $\texttt{num\_bins}=11$, $\texttt{alpha\_cls}=0.2$, $\texttt{pred\_weight}=0.5$.
\end{table*}


\begin{table*}[h!]
\centering
\small
\caption{\textbf{Ensemble} results on the Laptop Dev set for SemEval Task~1 (Valence/Arousal). We compare simple averaging, weighted averaging, and ridge-regression stacking (out-of-fold, OOF) with and without additional VADER-based features. Best (lowest) RMSE among valid (non-leaking) settings is highlighted.}
\label{tab:ensemble_results}
\begin{tabular}{llccccccc}
\hline
Setting & Method & MSE$\downarrow$ & RMSE$\downarrow$ & RMSE$_v$$\downarrow$ & RMSE$_a$$\downarrow$ & $\rho_v$$\uparrow$ & $\rho_a$$\uparrow$ & $\rho_{mean}$$\uparrow$ \\
\hline
\multirow{3}{*}{Without VADER}
& Avg & 0.4095 & 0.6399 & 0.5835 & 0.6918 & 0.9391 & 0.6105 & 0.7748 \\
& Weighted (w=[0.3,0.7]) & 0.4025 & \textbf{0.6344} & 0.5629 & 0.6987 & 0.9421 & 0.6124 & 0.7773 \\
& Stacking (Ridge, OOF) & 0.4025 & \textbf{0.6344} & 0.5713 & 0.6918 & 0.9397 & 0.5893 & 0.7645 \\
\hline
\multirow{3}{*}{With VADER}
& Avg & 0.4095 & 0.6399 & 0.5835 & 0.6918 & 0.9391 & 0.6105 & 0.7748 \\
& Weighted (w=[0.3,0.7]) & 0.4025 & \textbf{0.6344} & 0.5629 & 0.6987 & 0.9421 & 0.6124 & 0.7773 \\
& Stacking (Ridge, OOF) & 0.4079 & 0.6387 & 0.5718 & 0.6992 & 0.9396 & 0.5793 & 0.7594 \\
\hline
\end{tabular}
\end{table*}

\begin{table*}[h!]
\centering
\small
\setlength{\tabcolsep}{5pt}
\caption{Track~B (English) Environmental Protection --- best hybrid configuration and DEV results.}
\label{tab:trackb_envprot_best}
\begin{tabular}{lccc}
\hline
\textbf{Best config} & \textbf{num\_bins} & \textbf{$\alpha_{\text{cls}}$} & \textbf{pred\_weight ($w$)} \\
\hline
 & 21 & 1.0 & 0.3 \\
\hline
\end{tabular}

\vspace{2mm}

\begin{tabular}{lcccccc}
\hline
\textbf{Variant} &
\textbf{MSE} &
\textbf{RMSE} &
\textbf{RMSE$_v$} &
\textbf{RMSE$_a$} &
\textbf{Pearson$_v$} &
\textbf{Pearson$_a$} \\
\hline
Regression
& 2.0287 & 1.4243 & 1.7162 & 1.0546 & 0.5311 & 0.2221 \\
Bin-Expected
& 2.2221 & 1.4907 & 1.8301 & 1.0464 & 0.5272 & 0.0664 \\
Average ($w{=}0.3$)
& 1.9661 & 1.4022 & 1.6932 & 1.0322 & 0.5312 & 0.2234 \\
\hline
\end{tabular}

\vspace{2mm}
\footnotesize{\textit{Note.} Pearson mean: Regression = 0.3766, Bin-Expected = 0.2968, Average = 0.3773. Also: $\mathrm{RMSE}_{\text{avg}}=1.4022$, $\mathrm{RMSE}_{\text{reg}}=1.4243$, $\mathrm{RMSE}_{\text{cls}}=1.4907$.}

\vspace{2mm}
\footnotesize
$^\dagger$Full-fit stacking is trained on the full dev set and evaluated on the same dev set; it is therefore optimistic and should not be used for fair model selection. OOF stacking is the appropriate estimate.
\end{table*}

\begin{table}[t]
\centering
\setlength{\tabcolsep}{3pt}
\footnotesize
\begin{tabular}{lccccc}
\toprule
Model & MSE & RMSE & RMSE$_v$ & RMSE$_a$ & $\rho$ \\
\midrule
LLM (ICL) & 0.484 & 0.695 & 0.633 & 0.752 & 0.757 \\
\bottomrule
\end{tabular}
\caption{LLM dev results (Laptop).}
\label{tab:llm-dev}
\end{table}
\subsection{Ensemble on TrackA Laptop Dev}
Table \ref{tab:ensemble_results} shows the results from ensemble learning of two models, with/without the VADER feature on the \textit{laptop} data, which show much improvement in comparison to individual models (hybrid RoBERTa and LLMs), especially on RMSE scores.
For the weighted ensemble, weights were selected via grid search over the interval [0,1] with step size 0.1, optimizing RMSE on the development set.
The goal is that 1) better models get more influence; 2) it reduces bias compared to equal averaging.
The experimental results show that average weighting produced RMSE score 0.6399 vs Weighted 0.6344 (lower and better).
The results also show that the VADER feature did not make an improvement in the evaluation scores, even a slight degradation, which indicates that VADER probably adds noise or a redundant signal.
In addition, the weighted avg and stacking produced the same RMSE scores (0.6344), although other scores are different.
This might suggest that the linear ridge basically learned weights approximately [0.3, 0.7], so stacking approximately a weighted average.
This often happens with only 2 models. To explore this, future work shall explore more models for the ensemble.


% ...




\subsection{Hybrid RoBERTa on Track-B}
Table \ref{tab:trackb_envprot_best} shows the performance of Hybrid RoBERTa on Track-B English data - Environmental Protection. The best hybrid configuration and the evaluation metrics are listed. Once again, the hybrid model performed the best over individual regression and bin-expected models. 

\section{Model Submissions}
Due to unforeseen situations, we were only available to submit the system output for TaskA.1 using the Hybrid RoBERTa model, i.e., without the LLMs and ensemble-learning variations. In addition, we did not submit for Track-B.
The initial/unofficial ranking from the organisers on our Hybrid ReBERTa (lightweight) shows that it achieved 16/30 and 22/33 on laptop and restaurant data, respectively. 
On laptop data, the best performing team has a score of 1,2408, while Hybrid RoBERTa has \textbf{1,4062}, which is closer to the best team and much better than the bottom-ranking team 1,8486 and baseline 2,8053.
Similarly, on restaurant data, the best performing team has 1,1035 error score, while Hybrid RoBERTa produced \textbf{1,3632}, much better than the last team 1,9115 and baseline 2,791.
Considering the very \textbf{low cost} from encoder-based Hybrid RoBERTa with \textbf{constrained} training, this performance is very promising, as shown in Figure \ref{fig:eng-lap-A1} and \ref{fig:eng-res-A1} (Appendix).

\section{Conclusions and Future Work}

In this system paper, we introduced a hybrid encoder model RoBERTa averaging the weighted performance of regression and discretized classification heads. In addition, we explored the prediction-level fusion (late fusion or model stacking) for ensembling the output scores of the hybrid encoder and LLM; however, for future work, we would like to explore different kinds of ensemble methods, e.g., stacking ensemble from \cite{romero-etal-2025-medication}, as well as automatic hyper-parameter finetuning \cite{akiba2019optunanextgenerationhyperparameteroptimization}.

\section*{Limitations}
Due to time limitations, we did not apply LLMs and ensembles on the test set, but on the dev set. We will explore our system performance on test sets offline. To test model generalisability, we plan to explore the performance on languages other than English, such as Chinese.


\bibliography{custom}

\clearpage

\appendix

\section{Example of Shared Task Data}
\begin{figure}[t]
  \includegraphics[width=0.5\textwidth]{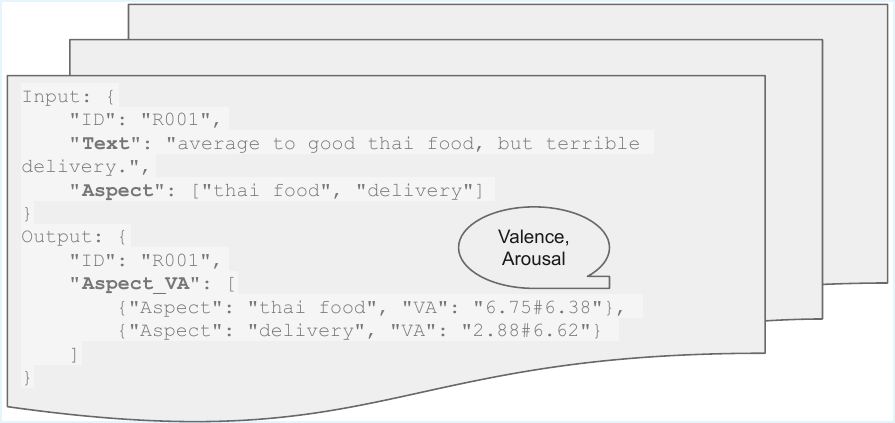}
  \caption{Example of TaskA1 data
  } 
  \label{fig:example_taskA1}
\end{figure}

\begin{figure}[t]
  \includegraphics[width=0.5\textwidth]{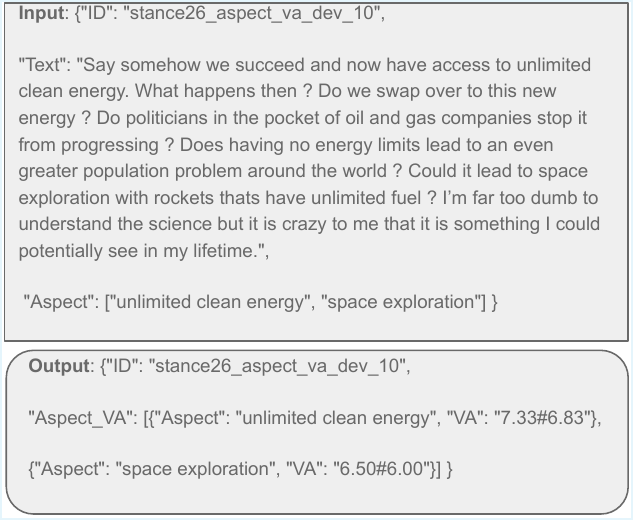}
  \caption{Example of TaskB data
  } 
  \label{fig:example_taskB1}
\end{figure}
Figure \ref{fig:example_taskA1} and \ref{fig:example_taskB1} display the example data of Track A and B.

\section{QuadAI Ranking Among Teams}
\label{sec:appendix}

\begin{figure}[t!]
  \centering
  \includegraphics[width=0.3\textwidth]{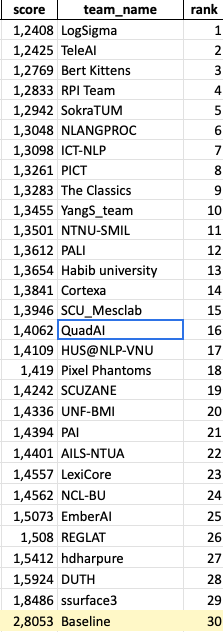}
  \caption{English Laptop Data Track-A1
  } 
  \label{fig:eng-lap-A1}
\end{figure}

\begin{figure}[t]
  \centering
  \includegraphics[width=0.3\textwidth]{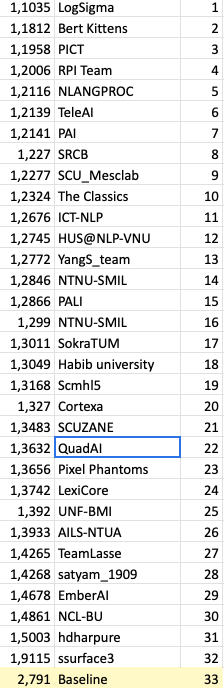}
  \caption{English Restaurant Data Track-A1
  } 
  \label{fig:eng-res-A1}
\end{figure}

We list the initial/un-official ranking we received for reference in Figure \ref{fig:eng-lap-A1} and \ref{fig:eng-res-A1} on Track-A1 laptop and restaurant data respectively.

\section{Data Cleaning Prompt}
\label{sec:appendix:prompt}

The data cleaning prompt includes:
Persona, 
Metric Definitions,
Cluster examples, and 
Returns, which are detailed below.

\begin{lstlisting}[language=Python, caption=Cluster Critique Prompt for VA-based Sentiment Analysis]
CLUSTER_CRITIQUE_PROMPT = """You are an expert in sentiment analysis. 
Below is a cluster of aspect-sentiment examples grouped by similar 
Valence-Arousal (VA) values.

DEFINITIONS:
- Valence: 1=very negative to 9=very positive
- Arousal: 1=very calm to 9=very intense

CLUSTER EXAMPLES:
{examples_text}

TASK: Identify which examples (if any) have spurious/incorrect VA labels 
that don't match the text sentiment. An example is spurious if:
1. The VA values don't match the sentiment expressed in the text
2. The aspect sentiment is clearly different from the labeled values
3. The label seems inconsistent with similar examples in this cluster

Return ONLY valid JSON with NO extra text:
{{
  "spurious_indices": [0, 5, 12],
  "reasoning": "Example 0: Text is very negative but valence is too high. 
  Example 5: ..."
}}

If no examples are spurious, return: 
{{"spurious_indices": [], "reasoning": "All labels appear correct"}}"""
\end{lstlisting}

\end{document}